# PointGrasp: Point Cloud-based Grasping for Tendon-driven Soft Robotic Glove Applications*

Chen Hu, Shirui Lyu, Eojin Rho, Daekyum Kim, Shan Luo and Letizia Gionfrida

*Abstract*— Controlling hand exoskeletons to assist individuals with grasping tasks poses a challenge due to the difficulty in understanding user intentions. We propose that most daily grasping tasks during activities of daily living (ADL) can be deduced by analyzing object geometries (simple and complex) from 3D point clouds. The study introduces *PointGrasp*, a real-time system designed for identifying household scenes semantically, aiming to support and enhance assistance during ADL for tailored end-to-end grasping tasks. The system comprises an RGB-D camera with an inertial measurement unit, and a microprocessor integrated into a tendon-driven soft robotic glove. The RGB-D camera process 3D scenes at a rate exceeding 30 frames per second. The proposed pipeline demonstrates an average RMSE of 0.8 ± 0.39 cm for simple and of 0.11 ± 0.06 cm for complex geometries. Within each mode, it identifies and pinpoints reachable objects. This system shows promise in end-to-end vision-driven robotic-assisted rehabilitation manual tasks. The code is released at https://vhr.github.io/.

## I. INTRODUCTION

Over the past decade, wearable robotic hand orthoses have been developed and tested for individuals with motor hand disorders [1]. These systems have shown the potential to enhance traditional rehabilitation therapy programs and assist in activities of daily living (ADL) by providing external forces to the hands [2]. Recent advancements in wearable robotic hand orthoses have culminated in pneumatic gloves [3] and linkage-based or cable-driven systems [4]. Cable-driven gloves, on the other hand, have shown precision and force [5], making them more suitable for tasks requiring fine control. In both scenarios, a critical concern involves determining how to effectively regulate the wearable robotic hand orthoses based on the movement intentions of the wearer [6]. Although both approaches are valid, we focus on underactuated tendon-driven gloves.

*This work was supported by NMES PhD studentships provided by King's College London.

Chen Hu, Shirui Lyu and Letizia Gionfrida are with the Department of Informatics, Faculty of Natural Mathematical and Engineering Sciences, King's College London, Bush House, London, WC2R 2LS, United Kingdom (email: {tyrone.hu, shirui.lyu, letizia.gionfrida}@kcl.ac.uk).

Eojin Roh is with School of Computing, KAIST, Daejeon 34141, South Korea (email: djwls9453@kaist.ac.kr).

Daekyum Kim is with School of Mechanical Engineering and the School of Smart Mobility, Korea University, Seoul 02841, South Korea (email: daekyum@korea.ac.kr).

Shan Luo is with the Department of Engineering, Faculty of Natural Mathematical and Engineering Sciences, King's College London, London, WC2R 2LS, United Kingdom (email: shan.luo@kcl.ac.uk).

Letizia Gionfrida is with the John A. Paulson School of Engineering and Applied Sciences, Harvard University, Cambridge, MA, USA (email: gionfrida@seas.harvard.edu).

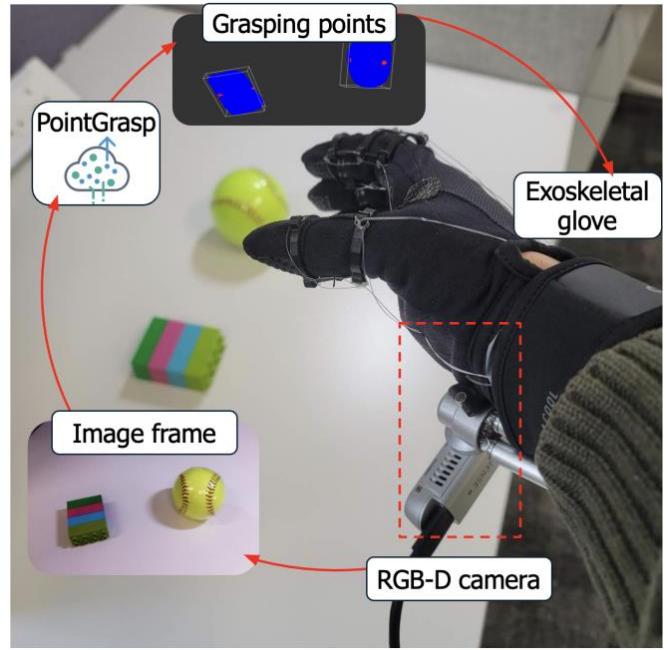

**Fig. 1:** Overview of the system depicting an RGB-D camera integrated into a tendon-driven soft wearable robotic glove developed for the study. Captured image frames undergo processing with the proposed *PointGrasp* algorithm to extract grasping points, which are fed back into the soft wearable robotic glove system.

While electromyography-based approaches (EMG), control methods based on neuromuscular interfaces, are effective [7], [8], these approaches are not able to perceive environmental features like object sizes and shapes.

To overcome these limitations, the exoskeleton community has explored the integration of vision-based environmental cues to enhance hand orthoses controllers. Existing systems predominantly rely on cameras integrated into glasses and a wearable hand orthosis worn by the user [9]. However, first-person view (FPV) cameras suffer from occlusion when the hand is approaching an object [10]. Moreover, current approaches advocate for vision solutions that offer high-level of assistance based on classes of visual inputs [11].

Existing vision-based approaches have not addressed the critical aspect of determining where to grasp. This is a crucial factor for achieving successful grasps because it affects the grasp's stability, efficiency, and functionality, influencing the overall effectiveness of the robotic manipulation tasks. The field of robotic grippers has successfully employed geometric point clouds to detect the presence of grasp points [12], [13]. While this approach has not been explored previously, integrating insights from this domain holds the potential to offer valuable perspectives for enhancing the functionality of vision-based wearable glove controllers.

We hypothesized that the grasping strategy between a robotic gripper and a wearable robot glove would be similar. This is because most existing tendon-based wearable gloves are based on under-actuation, and they move the index and middle fingers together to the thumbs, resembling the grasping pattern of a robot gripper. Despite its apparent straightforward nature, to the best of our knowledge, this approach has not been explored in existing literature. Based on the hypothesis we believe that there is an opportunity to directly detect grasping points in real-time, to present a solution where vision can be offered a tailored control strategy for both high-level and low-level assistance.

This work introduces *PointGrasp*, a real-time grasping detector system. To achieve this, a wearable robotic glove (Fig. 1) that uses tendon actuation was implemented for the study. To reduce FPV occlusion, the proposed approach, incorporates a wrist-mouted RGB-D camera into the developed wearable robotic glove. The algorithm we propose identifies grasping points, prompting the fingers to assume a grasping pose, thereby presenting a proof-of-concept application of a vision-based controller that extends beyond high-level functions.

II. METHODS

The proposed approach consisted of an RGB-D camera for sensing environmental geometry integrated with a wearable robotic glove. The proposed *PointGrasp* (Fig. 2) is designed to sense the object geometry (basic or complex), for complex geometries the system segments the handles and outputs the grasping points. The proposed approach operates at real-time frame rate (≥30 fps) and at an 800x600 resolution to reduce computational costs. Finally, *PointGrasp* presents a semantic label to guide further scene parsing to detect objects and identify potential grasp heuristics.

*A. Dataset*

We employed a diverse set of objects from the Yale-CMU-Berkeley (YCB) dataset [14]. The YCB set consists of 77 objects of daily life with different shapes, sizes, textures, weight, and rigidity, as well as some widely used manipulation tests. We adopted 18 commonly used [15] YCB objects for the experiments of the YCB datasets, aimed aiming to achieve a comprehensive and representative sample of the overall set. The left image in Fig. 2 illustrates the array of 9 YCB objects with simple (or primitive) geometries (Fig. 3) and 9 YCB objects with complex geometries (Fig. 4) that constitute the dataset on which the proposed *PointGrasp* was tested. We collected 400 RGB-D frames from simple geometries and 400 frames from complex geometries.

*B. Hardware set-up*

In this study we designed a wearable robotic glove (Fig. 1). The system included an actuation mechanism connected to the glove via a Tendon-Sheath Mechanism (TSM), ensuring minimal dynamic sheath bending angle changes. The glove was crafted from flexible polyamide, with 3D printed rings near the metacarpophalangeal, proximal interphalangeal, and distal interphalangeal joints, replicating human hand tendons for effective grasping and releasing motions. Actuation was driven by forearm-mounted DC motors, positioned to work in opposition, facilitating the glove's grasping and releasing. The motors operated by alternately pulling flexion wires and releasing extension wires, or vice versa, to manipulate the glove. The glove follows the design proposed by Rho et al.

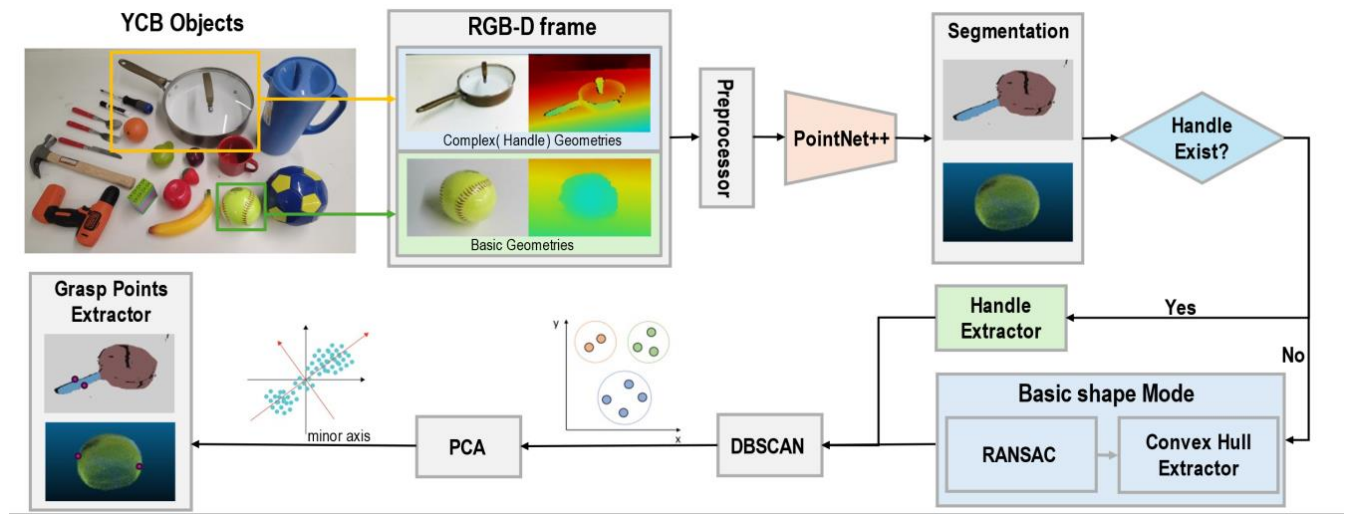

Fig. 2: Illustration showing the *PointGrasp* architecture for real-time 3D object sensing. On the left, YCB objects are captured by RGB-D cameras, with simple geometries enclosed in the green box and complex geometries in the orange box. Point cloud data is extracted using PointNet++ [1]. In the absence of handles, Random Sample Consensus (RANSAC) [20] is applied, followed by convex hull extraction. For complex geometries, a classifier identifies handles. The data undergo denoising with Density-Based Spatial Clustering of Applications with Noise (DBSCAN) [21] and further processing with Principal Component Analysis (PCA) to extract the point cloud of target objects, represented by red points in the bottom-left image.

[16]. To sense the environment, the wearable robotic glove was combined with a wrist-mounted Intel RealSense D435i.

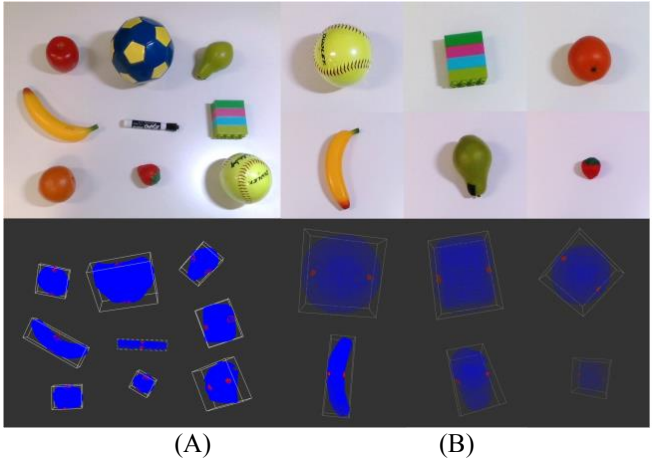

**Fig. 3:** Illustration showing *PointGrasp* output for 9 YCB objects with simple geometries for multiple objects (A) and single objects (B) and their respective grasping points (red) in output.

### C. PointGrasp architecture

The proposed *PointGrasp* architecture identifies the grasping points under two different modes. The first mode includes basic (or primitive) geometries. The basic models are cuboid, a sphere, etc. The second mode identifies components of the basic geometries, such as handles and include objects like skillet, pitcher, mug, etc.

The initial mode comprises fundamental or primitive geometries, including shapes like cuboids and spheres. In this mode, the center of mass aligns with the grasping points on the object. The second mode involves the identification of specific components within these basic geometries, such as handles. Objects falling under this category, such as skillets, pitchers, mugs, etc., deviate from the alignment of the center of mass with the location where the object is being grasped. These modes include a sufficient variety of YCB objects for grasping tasks.

Given the focus of the proposed work on three-dimensional scenes, computational efficiency becomes a crucial consideration. To address this concern, we opted to leverage the open-source library Open3D [17] along with the point cloud format for the representation of 3D images. The utilization of a point cloud format, which presents a 3D image as an array of points rather than a two-dimensional array of z-coordinates, aligns with the requirements of three-dimensional operations, including but not limited to nearest neighbor detection, surface normal estimation, and the identification of three-dimensional surfaces and shapes. Moreover, Open3D offered us a standardized and optimized suite of cutting-edge functions tailored for the manipulation of point clouds, further enhancing the overall efficiency of our computational processes.

### 1) Labelling

RGB and depth images were transformed into point clouds through their integration with intrinsic camera parameters. Our dataset encompasses point clouds generated from the selected 18 YCB from distinct scenes, each featuring 40-point clouds captured from varied views. Of these, 9 scenes included objects with handles, while the remaining 9 were composed of objects with primary geometric shapes. To segment handles on objects, we meticulously annotated the targets with handles, locating 30-point clouds for training, validation, and testing, across different objects and viewpoints. In each point cloud image representation, a total of three category were identified. The main bodies of the 9 YCB objects with complex geometries, without their handles, were represented one category including skillet lid, pitcher base, mug, power drill, hammer, Phillips screwdriver, large marker, fork, spoon, and knife. The handles of the 9 YCB objects were labelled into as a separate category. The remaining portions including the background were labeled as a third category.

### 2) PointNet++ training and validation

The pre-trained PointNet++ [18], tailored for the purpose of this approach was adopted for the classification of the objects geometries and the identification of the handles.

An initial preprocessing steps was implemented to remove the pixels flagged by the camera as inaccurate, saturated, or inconsistent. Since the objects to be grasped did not appear at the edge of the camera view, a tailored masking algorithm was designed to remove the noises at the edge of the image frames. After converting each valid pixel to a 3D point, they were stored in a single-point cloud $P \in R^{3 \times N}$, where $N$ represents the number of valid points in the depth image.

We fine-tuned PointNet++ using our labeled dataset to improve handle segmentation accuracy for the 9 YCB objects with complex geometries. This allowed PointNet++ to segment handles with varying degrees of complexity. The goal of PointNet++ was to calculate the category score vector $c^{point} = \{c_i^{point} | i \in |1, m|\}$ of point cloud within a given candidate grasps $g$. Then, we design a handle extractor to extract point clouds of handles, which can be expressed as:

$$P^{handles} = F(P, c^{point}, \emptyset) \qquad (1)$$

Where $P^{handles}$ is the cloud point of handles, and $\emptyset$ is the extractor parameter. This architecture includes 4 set abstraction layers (SA), 4 feature propagation layers (FP), and 2 convolutional layers. For all experiment we use batch size 32, Adam optimizer with learning rate of 0.001. Our architecture was based on PyTorch 1.11.0. Intersection over Union (*IoU*) was determined by comparing segmented outputs with detailed ground truth. Finally, we used the principal components analysis (PCA) to identify the principal axes of a handle, and estimate the bounding box for complex geometries, to find a minor axis that was parallel to the identified plane.

The proposed system used a simple heuristic that desirable grasp points are often located at the center of the sides of the handle [19]. The phase involves the calculation of vectors that extend from the handle's center of mass to all constituent points. Angles between these vectors and the minor axis are computed, after which the vector exhibiting the minimal angular deviation is selected. The point corresponding to this vector is deemed the preliminary grasping point. To ensure bilateral symmetry and enhance grasp stability, the algorithm computes a grasping point that is symmetric relative to the center of mass.

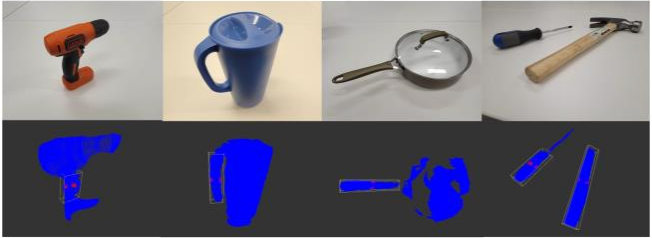

**Fig. 4:** Illustration showing *PointGrasp* output for 5 YCB objects with complex geometries. The top figure depicts the RGB image, and the bottom figure illustrates the handle identified and the grasp points (red) visualization of objects in the 3D box of the detected handles.

*3) Grasping points for intent detection*

If no handle existed in the point cloud, was used a random sample consensus (RANSAC) [20] for robust plane detection. RANSAC proved beneficial even when the depth image was cluttered, or the planes were obscured by other objects. Once a prominent plane of the object in the scene was identified, we identified the convex hull of the plane to delineate a defined surface. Subsequently, environmental noises were removed, and the convex hull was extruded along the normal surface to define the volume potentially containing the graspable object. Point clusters within the convex hull were then segmented to identify distinct objects. This process facilitated the identification of the principal axes of the object, the estimation of a three-dimensional bounding box, and the detection of the two grasping points (Fig. 4).

To give a measure of the accuracy of the grasp point detection, we compared the distance of each pair of estimated grasping points to the actual distance measurement of the two opposing centers of a rectangular object (a whiteboard eraser) along its minor axis using a caliper (0.25 mm accuracy). To evaluate the performance of the proposed approach in estimating grasping points on the 18 YCB with simple and complex geometries, we used the root-mean-square error (RMSE).

Once the grasping points were identified, the proposed architecture was able to infer the user's intent based on the orientation of the user's palm to actuate the fingers (Fig. 5). We computed the orthogonal Euclidean distance, denoted as $\vec{R}$ in Fig. 5, to each object within the scene. This approach ensured a vision-driven targeted interpretation of user intent, to trigger tailored actuation of the wearable robotic glove. As performance metrics, we calculated different hand-object distances to simulate a hand approaching an object for grasping and evaluated those in terms of time to completion and the travelling length.

## III. RESULTS

*1) Simple geometries*

Fig. 6 illustrates the grasp point accuracy for 9 YCB objects with simple geometries used in the study. These YCB objects include a banana, strawberry, apple, orange, marker, mini football, tennis ball, and Lego toy.

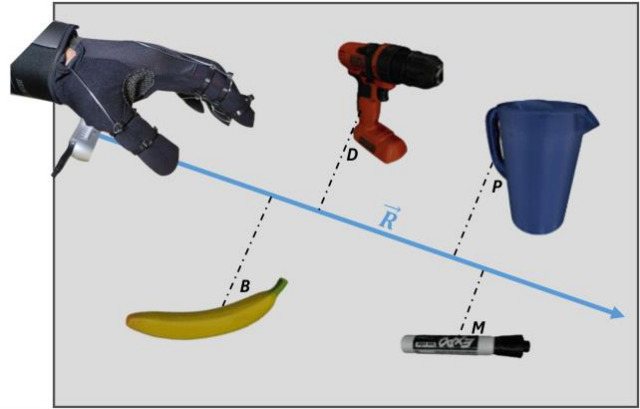

**Fig. 5:** Illustration of user intent prediction, leveraging the Euclidean distance from identified YCB objects (*B*: banana, *D*: drill, *M*: marker, and *P*: pitcher) to a vector $\vec{R}$, depicted in blue that goes from the camera situated beneath the user's palm within the exoskeleton glove.

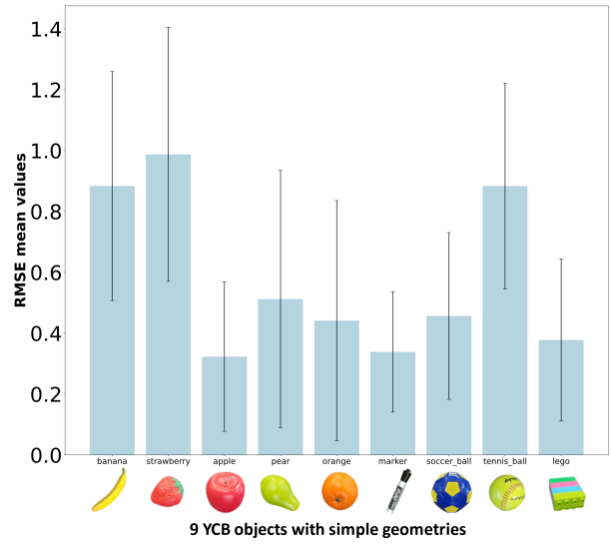

**Fig. 6:** RMSE of 9 YCB objects with simple geometries in grasp point distance estimation with their standard deviation.

The variability in the volume of these objects resulted in notable differences in the quantity of point clouds, impacting their operational range. Objects with more regular geometric shapes, such as apples, oranges, mini soccer balls, tennis balls, big markers, and Legos, consistently maintain an error below 1 cm within the optimal operational range. This stability persists despite minor fluctuations caused by motion and lighting. In the case of less regular geometric shapes, such as bananas and strawberries, the proposed method achieved errors below 1.4 cm and 0.8 cm within the optimal operational range, respectively. Notably, our approach excels in capturing the dimensions of smaller objects, equal to or less than 3.5 cm, including big markers (actual distance approximately 1.7 cm) and strawberries (actual distance around 3.6 cm).

*2) Complex geometries*

Fig. 7 illustrates the RMSE for 9 YCB objects with complex geometries, where the handles were present, including skillet, pitcher, mug, power drill, hammer, screw, fork, spoon, and knife. As the handle detection was based on was determined by a re-trained pre-trained PointNet++, Table I presents the IoU metrics for the tested objects, calculated by comparing the segmentation results against finely annotated ground truth data.

TABLE I.  IoU SEGMENTED OUTPUTS AGAINST GROUND TRUTH

| YCB Object/Handle | IoU |
|---|---|
| handle | 0.576 |
| Pitcher base | 0.886 |
| Mug | 0.727 |
| Skillet lid | 0.886 |
| Fork | 0.307 |
| Spoon | 0.259 |
| Knife | 0.255 |
| Power drill | 0.861 |
| Phillips screwdriver | 0.424 |
| Hammer | 0.525 |

We observed that the handles of these objects were influenced by factors such as illumination, camera view, distance, as well as their shape and volume. For instance, objects with curved handles (e.g., mug) or those with narrow handle widths (like a spoon measuring only 1.48 cm) impacted the accuracy of grasp point detection. On the other hand, an increase in handle volume did not enhance grasp point precision. This phenomenon can be attributed to the limited number of point clouds visible from a certain camera view, suggesting that larger volumes may result in increased self-occlusion. The consequent lack of point cloud data on the object's rear led to slightly higher error rates.

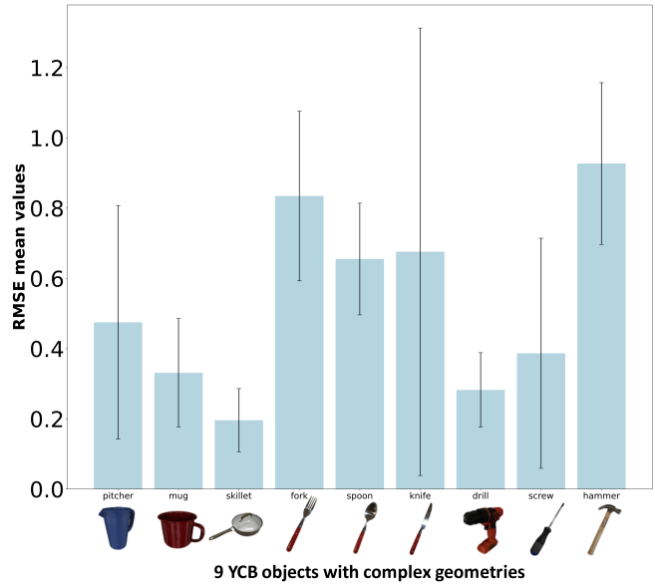

**Fig. 7:** RMSE of 9 YCB objects with complex geometries in grasp point distance estimation with their standard deviation.

*3) Grasp points for intent detection*

Once the grasp points of user's intent were detected, the fingers triggered into a grasping pose after approximately 3 seconds at an average distance of 30cm across the different YCB for objects (Fig. 8).

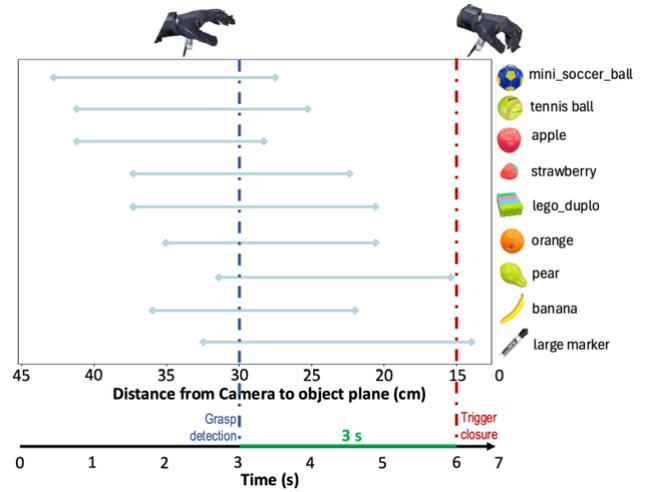

**Fig. 8:** Intent detection of the wearable robotic glove. The light blue lines illustrate operable range of different YCB objects. Once the grasping points are detected at around 30 cm (blue vertical line). Then the system triggers closure after 3 s (red vertical line).

## IV. DISCUSSION

The objective of this study is to introduce *PointGrasp* a method to extract graspoing points from simple and complex geometries for a wearable robotic glove with a wrist-mounted RGB-D camera.

In introducing our initial proof-of-concept system, we have demonstrated the system's capacity for real-time grasping. Although a simple grasping heuristic was used, the system proved to perform well at estimating grasping points on a variety of YCB objects. The approach of initially detecting the handles and then localizing the grasp points, aligns with human manipulation habits, as grasping an object by its handle typically precedes various subsequent actions. This ability empowers users to control a grasping-assistance solely using vision, exemplified by the system inferring the user's intent to grasp an object on a flat surface and triggering finger closure when the hand is positioned in the direction of the users' palm. This prototype holds promise for Activities of Daily Living (ADLs) tasks, particularly for individuals with reduced muscle activity, such as post-stroke survivors. By leveraging environmental cues, the proposed *PointGrasp* system reduces the need for external devices (e.g., camera into glassess).

While proposed model correctly identifies the region of correct grasping, it currently lacks an evaluation of how user grasp, in terms of stability, with and without the vision-driven control, align with the identified grasping points. In particular, this proof-of-concept evaluation falls short in providing a success rate for grasping based on the visually-identified points. This crucial aspect will undergo thorough examination in future studies.

In the results section, the focus on the proportion of correctly identified handle point clouds demonstrates a fast and efficient 3D point identification across different object shapes. The observed variation in accuracy across diverse objects offers insights into the model's sensitivity to handle shape complexity. In Fig. 6 and 7 we observed that larger objects are more susceptible to self-occlusion due to a single camera perspective, which significantly impacts the determination of grasping points. Future enhancements include the incorporation of a broader semantic label library, facilitating refined environmental geometry-based category identification.

The absence of user study across diverse subjetcs and manipulation tasks, will also consitute future investigations. The goal is to broaden the system's adaptability to diverse environmental contexts and household tasks. Future studies will also investigate and address needed improvements linked to depth camera noise by incorporating advanced filtering for planar identification, essential to mitigate the impact of camera noise and enhance real-time precision in recognizing flat surfaces or planes within the 3D space. Future focus will extend to explore frame-to-frame registration, leveraging sensor histories to fine-tune the segmentation algorithm, and enhancing the precision of grasping points.

## V. Conclusion

The proposed system identifies grasping points of objects for an RGB-D camera positioned on the user's wrist. The model sets itself apart by being deterministic and predictive. To foster collaboration and adoption, the full pipeline code of the proposed *PointGrasp* is openly shared, enabling the community to seamlessly integrate the method into their pipelines.